\documentclass[runningheads]{llncs}
\usepackage{array}
\newcolumntype{P}[1]{>{\centering\arraybackslash}p{#1}}
\newcolumntype{M}[1]{>{\centering\arraybackslash}m{#1}}

\usepackage{hyperref}
\usepackage{amssymb}
\usepackage{caption}
\usepackage{booktabs}
\usepackage{amsmath}
\usepackage{subfigure}
\usepackage{graphicx}
\graphicspath{ {./images/} }
\usepackage{algorithm}
\usepackage{algpseudocode}
\algnewcommand\algorithmicforeach{\textbf{for each}}
\algdef{S}[FOR]{ForEach}[1]{\algorithmicforeach\ #1\ \algorithmicdo}
\begin{document}
\title{Speeding Up Recommender Systems Using Association Rules \thanks{Research co-funded by Polish National Science Centre (NCN) grant no. 2018/31/N/ST6/00610.}}
\author{Eyad Kannout\inst{1,2}\orcidID{0000-0001-7543-774X} \\ 
Hung Son Nguyen\inst{1,3}\orcidID{0000-0002-3236-5456} \\
Marek Grzegorowski\inst{1,4}\orcidID{0000-0003-4740-0725}}
\authorrunning{E. Kannout Author et al.}
\institute{Institute of Informatics, University of Warsaw, Warsaw, Poland \and
\email{eyad.kannout@mimuw.edu.pl} \and
\email{son@mimuw.edu.pl} \and
\email{m.grzegorowski@mimuw.edu.pl}
}
\maketitle              % typeset the header of the contribution
\begin{abstract}
% The abstract should briefly summarize the contents of the paper in
% 15--250 words.
Recommender systems are considered one of the most rapidly growing branches of Artificial Intelligence. The demand for finding more efficient techniques to generate recommendations becomes urgent. However, many recommendations become useless if there is a delay in generating and showing them to the user. Therefore, we focus on improving the speed of recommendation systems without impacting the accuracy.

\indent In this paper, we suggest a novel recommender system based on Factorization Machines and Association Rules (FMAR). We introduce an approach to generate association rules using two algorithms: (i) apriori and (ii) frequent pattern (FP) growth. These association rules will be utilized to reduce the number of items passed to the factorization machines recommendation model. We show that FMAR has significantly decreased the number of new items that the recommender system has to predict and hence, decreased the required time for generating the recommendations. On the other hand, while building the FMAR tool, we concentrate on making a balance between prediction time and accuracy of generated recommendations to ensure that the accuracy is not significantly impacted compared to the accuracy of using factorization machines without association rules.

\keywords{Recommendation system \and Association rules \and Apriori algorithm \and Frequent pattern growth algorithm \and Factorization machines \and Prediction’s time \and Quality of recommendations.}
\end{abstract}
\section{Introduction}
Throughout \footnote[1]{The final publication is available at Springer via https://doi.org/10.1007/978-3-031-21967-2\_14} the past decade, recommender systems have become an essential feature in our digital world due to their great help in guiding the users towards the most likely items they might like. Recently, recommendation systems have taken more and more place in our lives, especially during the COVID-19 pandemic, where many people all over the world switched to online services to reduce the direct interaction between each other. Many researchers do not expect life to return to normal even after the epidemic. All previous factors made recommender systems inevitable in our daily online journeys.

%  are leveraging the historical data and predicting users’ interests
% this idea can be used with any recommender system, not only factorization machine

\indent Many online services are trying to boost their sales by implementing recommendation systems that estimate users' preferences or ratings to generate personalized offers and thus recommend items that are interesting for the users. Recommendation systems can be built using different techniques which leverage the rating history and possibly some other information, such as users' demographics and items' characteristics. The goal is to generate more relevant recommendations. However, these recommendations might become useless if the recommendation engine does not produce them in a proper time frame.

Recently, the factorization machine has become a prevalent technique in the context of recommender systems due to its capabilities of handling large, sparse datasets with many categorical features. Although many studies have proved that factorization machines can produce accurate predictions, we believe that the prediction time should also be considered while evaluating this technique. Therefore, in this paper, we work on finding a novel approach that incorporates association rules in generating the recommendations using the factorization machines algorithm to improve the efficiency of recommendation systems. It is worth noting that the factorization machine model is used to evaluate our method and compare the latency of FM before and after using the association rules. However, in practice, our method can be combined with any other recommendation engine to speed up its recommendations. 

\indent The main contributions of this paper are as follows: 1) proposing a method that uses the apriori algorithm or frequent pattern growth (FP-growth) algorithm to generate association rules which suggest items for every user based on the rating history of all users; 2) utilizing these association rules to create short-listed set of items that we need to generate predictions for them; 3) employing factorization machines model to predict missing user preferences for the short-listed set of items and evaluate the top-N produced predictions.

% producing new aggregated forms of user-item matrix based on previous grouping of ratings and users
% 

\indent The remainder of this paper is organized as follows. In Section \ref{sec:preliminaries}, we provide background information for factorization machines algorithm and association rules in addition to reviewing some related works. In Section \ref{sec:FMAR}, we describe the problem we study in this paper. Also, we present FMAR - a novel recommender system that utilizes factorization machines and association rules to estimate users' ratings for new items. Section \ref{sec:eval} evaluates and compares FMAR with a traditional recommender system built without employing association rules. Finally, in Section \ref{sec:summary}, we conclude the study and suggest possible future work.

%\section{LITERATURE REVIEW FOR FACTORIZATION MACHINE AND ASSOCIATION RULES}
\section{Preliminaries}\label{sec:preliminaries}
In this section, we briefly summarize the academic knowledge of factorization machines and association rules.

\subsection{Factorization Machines}
In linear models, the effect of one feature depends on its value. While in polynomial models, the effect of one feature depends on the value of the other features. Factorization machines \cite{c1} can be seen as an extension of a linear model which efficiently incorporates information about features interactions, or it can be considered equivalent to polynomial regression models where the interactions among features are taken into account by replacing the model parameters with factorized interaction parameters. \cite{c2}.

However, polynomial regression is prone to overfitting due to a large number of parameters in the model. Needless to say, it is computationally inefficient to compute weights for each interaction since the number of pairwise interactions scales quadratically with the number of features. On the other hand, factorization machines elegantly handled previous issues by finding a one-dimensional vector of size k for each feature. Then, the weight values of any combination of two features can be represented by the inner product of the corresponding features vectors. Therefore, factorization machines manage to factorize the interactions weight matrix $W\in R^{n \times n}$, which is used in polynomial regression, as a product $VV^{T}$, where $V \in R^{n \times k}$. So, instead of modeling all interactions between pairs of features by independent parameters like in polynomial regression (see Equation 1). We can achieve that using factorized interaction parameters, also known as latent vectors, in factorization machines (see Equation 2).

\begin{equation}
\widehat{y}(x) = w_0 + \sum_{i=1}^{N} w_i x_i + \sum_{i=1}^{N} \sum_{j=i+1}^{N} w_{ij} x_i x_j
\end{equation}
% Where:\\
$ w_0\in\mathbb{R}$: is the global bias.\\
$ w_i\in\mathbb{R}^{n}$: models the strength of the i-th variable.\\
$w_{ij}\in\mathbb{R}^{n\times n}$: models the interaction between the ith and j-th variable.

\begin{equation}
\widehat{y}(x) = w_0 + \sum_{i=1}^{N} w_i x_i + \sum_{i=1}^{N} \sum_{j=i+1}^{N} \langle v_i,v_j \rangle x_i x_j
\end{equation}
% $V \in \mathbb{R}^{n\times k}$.\\
$\langle v_i,v_j \rangle$: models the interaction between the i-th and j-th variable by factorizing it, where $V \in \mathbb{R}^{n\times k}$ and $\langle .,. \rangle$ is the dot product of two vectors of size k.\\

This advantage is very useful in recommendation systems since the datasets are mostly sparse, and this will adversely affect the ability to learn the feature interactions matrix as it depends on the feature interactions being explicitly recorded in the available dataset.

\subsection{Association Rules}
The basic idea of association rules \cite{c3}\cite{c4} is to uncover all relationships between elements from massive databases. These relationships between the items are extracted using every distinct transaction. In other words, association rules try to find global or shared preferences across all users rather than finding an individual’s preference like in collaborative filtering-based recommender systems.

At a basic level, association rule mining \cite{c3}\cite{c4}\cite{c5} analyzes data for patterns or co-occurrences using machine learning models. An association rule consists of an antecedent, which is an item found within the data, and a consequent, which is an item found in combination with the antecedent. Various metrics, such as support, confidence, and lift, identify the most important relationships and calculate their strength. Support metric \cite{c3}\cite{c4}\cite{c5} is the measure that gives an idea of how frequent an itemset is in all transactions (see Equation 3). The itemset here includes all items in antecedent and consequent. On the other hand, the confidence \cite{c3}\cite{c4}\cite{c5} indicates how often the rule is true. In other words, it defines the percentage of occurrence of consequent given that the antecedents occur (see Equation 4). Finally, the lift \cite{c5} is used to discover and exclude the weak rules that have high confidence, which can be calculated by dividing the confidence by the unconditional probability of the consequent (see Equation 5). Various algorithms are in place to create associations rules using previous metrics, such as Apriori \cite{c4}\cite{c6}, AprioriTID \cite{c4}\cite{c6}, Apriori Hybrid \cite{c4}\cite{c6}, AIS (Artificial Immune System) \cite{c7}, SETM \cite{c8} and FP-growth (Frequent pattern) \cite{c4}\cite{c9}. In the next section, we provide more details about how we use these metrics to find the association rules used to improve the prediction time in the recommender system.

\begin{equation}
\label{support}
\small
\text{Support}(\{X\}\rightarrow\{Y\}) =  \frac{\text{Transactions containing both X and Y}}{\text{Total Number of transactions}}
\end{equation}

\begin{equation}
\label{confidence}
\small
\text{Confidence}(\{X\}\rightarrow\{Y\}) =  \frac{\text{Transactions containing both X and Y }} {\text{Transactions containing X}}
\end{equation}

\begin{equation}
\label{lift}
\small
\text{Lift}(\{X\}\rightarrow\{Y\}) =  \frac{\text{Confidence}}{\text{Transactions containing Y}}
\end{equation}

% %%%%%%%%%%%%%%%%%%%%%%%%%%%%%%%%%%%%%%%%%%%%%%%%%%%%%%%%%%%%%%%%%%%%%%%%%%%%%%%%

%\subsection{RELATED WORKS}
\subsection{Related works}
Over the past decade, a lot of algorithms concerned with improving the accuracy of the recommendation have been constantly proposed. However, while reviewing the research literature related to recommendation systems and what has been done to improve the prediction's time, we find that there is a research gap in this area even though the speed of recommendation, besides the accuracy, is a major factor in real-time recommender systems.
% contextual collaborative filtering approaches that utilize contextual information to improve recommendation quality

Xiao et al. \cite{c10} worked on increasing the speed of recommendation engines. They spotted that the dimension of the item vector in a collaborative filtering algorithm is usually very large when we calculate the similarity between two items. To solve this problem, they introduced some methods to create a set of expert users by selecting small parts of user data. One of these methods is based on selecting expert users according to the number of types of products they have purchased before. In comparison, another method calculates the similarities between users and then selects expert users based on the frequency that the user appears in other users’ K-most similar users set. The results show that using expert users in an item-based collaborative filtering algorithm has increased the speed of generating recommendations with preserving the accuracy to be very close to original results. Tapucu et al. \cite{c11} carried out some experiments to check the performance of user-based, item-based, and combined user/item-based collaborative filtering algorithms. Different aspects have been considered in their comparisons, such as size and sparsity of datasets, execution time, and k-neighborhood values. They concluded that the scalability and efficiency of collaborative filtering algorithms need to be improved. The existing algorithms can deal with thousands of users within a reasonable time. Still, modern e-commerce systems require to scale to millions of users and hence, expect even improved prediction time and throughput of recommendation engines. According to previous findings, we believe there is room for further improvements concerning prediction time and efficiency of recommendation systems.

% %%%%%%%%%%%%%%%%%%%%%%%%%%%%%%%%%%%%%%%%%%%%%%%%%%%%%%%%%%%%%%%%%%%%%%%%%%%%%%%%

%\section{FMAR RECOMMENDER SYSTEM}
\section{FMAR Recommender System \label{sec:FMAR}}
In this section, we formally provide the statement of the problem that we aim to tackle. Then, we introduce the details of a novel recommender system that is based on Factorization Machine and Association Rules (FMAR). We first formalize the problem we plan to solve. Then, we describe our proposed model which has two versions based on the algorithm used to generate the association rules: (i) factorization machine apriori based model, and (ii) factorization machine FP-growth based model.

% factorization machine apriori-based model that is used in FMAR.

\subsection{Problem Definition}
In many recommender systems, the elapsed time required to generate the recommendations is very crucial. Moreover, in some systems, any delay in generating the recommendations can be considered as a failure in the recommendation engine.
The main problem we address in this paper is to minimize the prediction latency of the recommender system by incorporating the association rules in the process of creating the recommender system. The main idea is to use the association rules to decrease the number of items that we need to approximate their ratings, hence, decreasing the time that the recommender system requires to generate the recommendations. Also, our goal is to make sure that the accuracy of the final recommendations is not impacted after filtering the items using the association rules.
% it is considered as important as the accuracy of recommendations because .

\subsection{Factorization Machine Apriori Based Model}
%  we need to explain here how do we extract association rules
In this section, we introduce the reader to the first version of FMAR which proposes a hybrid model that utilizes factorization machines \cite{c1} and apriori \cite{c4}\cite{c6} algorithms to speed up the process of generating the recommendations. Firstly, we use apriori algorithm to create a set of association rules based on the rating history of users. Secondly, we use these rules to create users' profile which recommends a set of items for every user. Then, when we need to generate recommendations for a user, we find all products that are not rated before by this user, and instead of generating predictions for all of them, we filter them using the items in the users' profile. Finally, we pass the short-listed set of items to a recommender system to estimate their ratings using factorization machines model.

In the context of association rules, it is worth noting that while generating the rules, all unique transactions from all users are studied as one group. On the other hand, while calculating the similarity matrix in collaborative filtering algorithms, we need to iterate over all users and every time we need to identify the similarity matrix using transactions corresponding to a specific user. % Future work, publication, this idea can help in cold start problem by using the generic rule that has no antecedents
However, what we need to do to improve the recommendation speed is to generate predictions for parts of items instead of doing that for all of them. Next, we introduce the algorithms that we use to generate the association rules and users' profile (cf. Algorithm \ref{alg:cap1}).
% The main steps of this algorithm are as follows:
% \begin{itemize}
% \item Create a dataset of only favorable reviews by filtering out the ratings which are less than a threshold.
% % \item Create a dictionary of high rated movies per user.
% \item Find frequent item-sets, according to equation ~\ref{support}, which contain items that have been highly rated together more than min\_support threshold.
% % % \item Find frequent itemset with length > 1
% % \item Find frequent itemset with length > 1. Here, in every frequent itemset all elements should have been highly rated together by users more than min\_support threshold.
% \item Extract all possible association rules from frequent item-sets where every rule has antecedent$>$=0 item, and consequent=1 item. 
% % which has the following structure:\\
% % premise: antecedent\\
% % conclusion: consequent (always contain one element)
% \item Compute confidence for every rule according to equation~\ref{confidence}.
% % using the formula: P(Y|X) = P(Y∧X) / P(X).
% \item Filter out the rules that have less than min\_confidence threshold.
% \item Create users' profile which recommends a set of items for every user as follows:
%     \begin{itemize}
%     \item For every user, find high rated items based on the rating history.
%     \item Find all association rules that their antecedents are subset of high rated items of that user.
%     \item Recommend all consequences of these rules to the user.
%     \end{itemize}
% \end{itemize}

\begin{algorithm}
\caption{Association Rules Generation Using Apriori Algorithm}\label{alg:cap1}
\begin{algorithmic}[1]
\State Extract favorable reviews \Comment{ratings $>$ 3}
\State Find frequent item-sets $\mathcal F$ \Comment{support $>$ min\_support}
\State Extract all possible association rules $\mathcal R$
% \ForEach {$f \in \mathcal F $}
% \State Extract all possible association rules $\mathcal R$
% \EndFor
\ForEach {$r \in \mathcal R $}
\State Compute confidence and lift
\If {(confidence $<$ min\_confidence) or (lift $<$ min\_lift)}
\State Filter out this rule from $\mathcal R$
\EndIf
\EndFor
\State Create users’ profile using rules in $\mathcal R$
\end{algorithmic}
\end{algorithm}

\vspace{-2em}

\begin{algorithm}
\caption{Users’ Profile Generation}\label{alg:cap2}
\begin{algorithmic}[1]
\ForEach {user}
\State Find high rated items based on rating history
\State Find the rules that their antecedents are subset of high rated items
\State Recommend all consequences of these rules 
\EndFor
\end{algorithmic}
\end{algorithm}

After generating the users' profile, we can use it to improve the recommendation speed for any user by generating predictions for a subset of not-rated items instead of doing that for all of them. The filtering is simply done using the recommended items which are extracted for every user using the association rules (cf. Algorithm \ref{alg:cap2}). Moreover, the filtering criteria can be enhanced by using the recommended items of the closest n-neighbors of the target user. The similarity between the users can be calculated using pearson correlation or cosine similarity measures.

On the other hand, it is noteworthy that the association rules in our experiments are generated using the entire dataset which means that these rules try to find global or shared preferences across all users. However, another way to generate the association rules is to split the dataset based on users' demographics, such as gender, or items' characteristics, such as genre, or even contextual information, such as weather and season. Thus, if we are producing recommendations for a female user in winter season, we can use dedicated rules which are extracted from historical ratings given by females in winter season. Following this strategy, we can generate multiple sets of recommendation rules which can be used later during prediction time to filter the items. Obviously, the rules generated after splitting the dataset will be smaller. So, the prediction latency can be minimized by selecting a smaller set of rules. In fact, this feature is very useful when we want to make a trade-off between the speed and quality of recommendations. Lastly, it is important to note that several experiments are conducted in order to select the appropriate values of hyper-parameters used in previous algorithms. For instance, min\_support $=250$, min\_confidence $=0.65$, number of epochs in FM $=100$, number of factors in FM $=8$. However, multiple factors are taken into consideration while selecting those values, including accuracy, number of generated rules, and memory consumption.

% \vspace{-0.1em}

\begin{algorithm}[H]
% \begin{algorithm}[b]
\caption{FP-Tree Construction}\label{alg:cap3}
\begin{algorithmic}[1]
% \ForEach {user}
% \State Find high rated items based on rating history
% \State Find the rules that their antecedents are subset of high rated items
\State Find the frequency of 1-itemset
\State Create the root of the tree (represented by null)
\ForEach {transaction}
\State Remove the items below min\_support threshold
\State Sort the items in frequency support descending order
\ForEach {item} \Comment{starting from highest frequency}
\If {item not exists in the branch}
\State Create new node with count 1
\Else
\State Share the same node and increment the count
\EndIf
\EndFor
\EndFor
% \EndFor
\end{algorithmic}
\end{algorithm}

\subsection{Factorization Machine FP-Growth based Model}
In this section, we introduce the second version of FMAR where FP-growth \cite{c4} \cite{c9} algorithm has been employed to generate the association rules. In general, FP-growth algorithm is considered as an improved version of apriori method which has two major shortcomings: (i) candidate generation of itemsets which could be extremely large, and (ii) computing support for all candidate itemsets which is computationally inefficient since it requires scanning the database many times. However, what makes FP-growth algorithm different from apriori algorithm is the fact that in FP-growth no candidate generation is required. This is achieved by using FP-tree (frequent pattern tree) data structure which stores all data in a concise and compact way. Moreover, once the FP-tree is constructed, we can directly use a recursive divide-and-conquer approach to efficiently mine the frequent itemsets without any need to scan the database over and over again. Next, we introduce the steps followed to mine the frequent itemsets using FP-growth algorithm. We will divide the algorithm into two stages: (i) FP-tree construction, and (ii) mine frequent itemsets (cf. Algorithm \ref{alg:cap3} and Algorithm \ref{alg:cap4}).

% the compressed representation of the database
% by traversing FP-tree we can efficiently  gre there’s no need to scan the database multiple times in order to calculate support. Instead, traversing the FP tree could do the same job more efficiently. since FP-growth can
% In general, FP-growth algorithm is designed to overcome the major shortcomings of  apriori algorithm by avoid the process of candidate generation that apriori relies on to find frequent itemsets. 

% \vspace{-2em}

\begin{algorithm}[H]
% \begin{algorithm}[b]
\caption{Mining Frequent Itemsets}\label{alg:cap4}
\begin{algorithmic}[1]
% \ForEach {user}
% \State Find high rated items based on rating history
% \State Find the rules that their antecedents are subset of high rated items
\State Sort 1-itemset in frequency support ascending order
\State Remove the items below min\_support threshold
\ForEach {1-itemset} \Comment{starting from lowest frequency}
\State Find conditional pattern base by traversing the paths in FP-tree
\State Construct conditional FP-tree from conditional pattern base
\State Generate frequent itemsets from conditional FP-ree
\EndFor
% \EndFor
\end{algorithmic}
\end{algorithm}

% \vspace{-1em}

After finding the frequent itemsets, we generate the association rules and users' profiles in the same way as in FM Apriori-based model. Regarding the hyper-parameters of FP-Growth algorithm, we used min\_support $=60$ and min\_confidence $=0.65$. Finally, in order to generate predictions, we employ a factorization machines model, which is created using the publicly available software tool libFM \cite{c12}. This library provides an implementation for factorization machines in addition to proposing three learning algorithms: stochastic gradient descent (SGD) \cite{c1}, alternating least-squares (ALS) \cite{c13}, and Markov chain Monte Carlo (MCMC) inference \cite{c14}.

% %%%%%%%%%%%%%%%%%%%%%%%%%%%%%%%%%%%%%%%%%%%%%%%%%%%%%%%%%%%%%%%%%%%%%%%%%%%%%%%%
\section{Evaluation for FMAR \label{sec:eval}}
%\section{EVALUATION FOR FMAR}
In this section, we conduct comprehensive experiments to evaluate the performance of the FMAR recommender system. In our experiments, we used MovieLens 100K dataset\footnote[1]{https://grouplens.org/datasets/movielens/} which was collected by the GroupLens research project at the University of Minnesota. MovieLens 100K is a stable benchmark dataset that consists of 1682 movies and 943 users who provide 100,000 ratings on a scale of 1 to 5. It is important to note that, in this paper, we are not concerned about users' demographics and contextual information since the association rules are generated based only on rating history.

% \subsection{Dataset}
% In our experiments, we used MovieLens 100K dataset\footnote[1]{https://grouplens.org/datasets/movielens/} which were collected by the GroupLens research project at university of Minnesota. MovieLens 100K is a stable benchmark dataset which consists of 1682 movies and 943 users who gave 100,000 ratings on a scale of 1 to 5. It is important to not that in this paper, we are not concerned about user demographics and contextual information since the association rules are created based only on the rating history. 

\subsection{Performance Comparison and Analysis}

In order to provide a fair comparison, we use several metrics and methods to evaluate FMAR and FM recommender systems, such as Mean Absolute Error (MAE), Normalized Discounted Cumulative Gain (NDCG), and Wilcoxon Rank-Sum test. Firstly, we selected 50 users who made a significant amount of ratings in the past. For every user, a dedicated testing set has been created by arbitrary selecting 70\% of the ratings made by this user in the past. On the other hand, the training set is constructed using the rest of the records in the entire dataset, which are not used in testing sets. This training set is used to generate the association rules and build the factorization machines model. For each evaluation method, we created two sets of items for every user. The first one, called \emph{original}, contains all items in the testing set. While the second one, called \emph{short-listed}, is created by filtering the original set using the association rules. 
% Then, we sort the results in descending order based on the predicted values, and find the top 10 items for every set.
% Finally, we 
Finally, we pass both sets to the factorization machines model to generate predictions and evaluate both versions of FMAR, i.e., Apriori-based and FP-growth-based FM models, by comparing them with the standard FM model operating on the complete data. 

% It is worth noting that the whole data set is used in all carried out experiments to generate the association rules and also to build the factorization machines model.
% There is no need to split the data into training and testing, because finally the predicted rating in both approaches (FM or FMAR) will be the same for the same item which is not filtered out by association rules. So, our approach is to check if FM is still able to predict something with high ratings in FMAR. In this way, we can confirm that the accuracy of FM is not impacted after using association rules.

\begin{figure}[t]
\centering
\begin{minipage}{.5\textwidth}
  \centering
  \captionsetup{justification=centering}
  \includegraphics[width=1\linewidth,height=4.0cm]{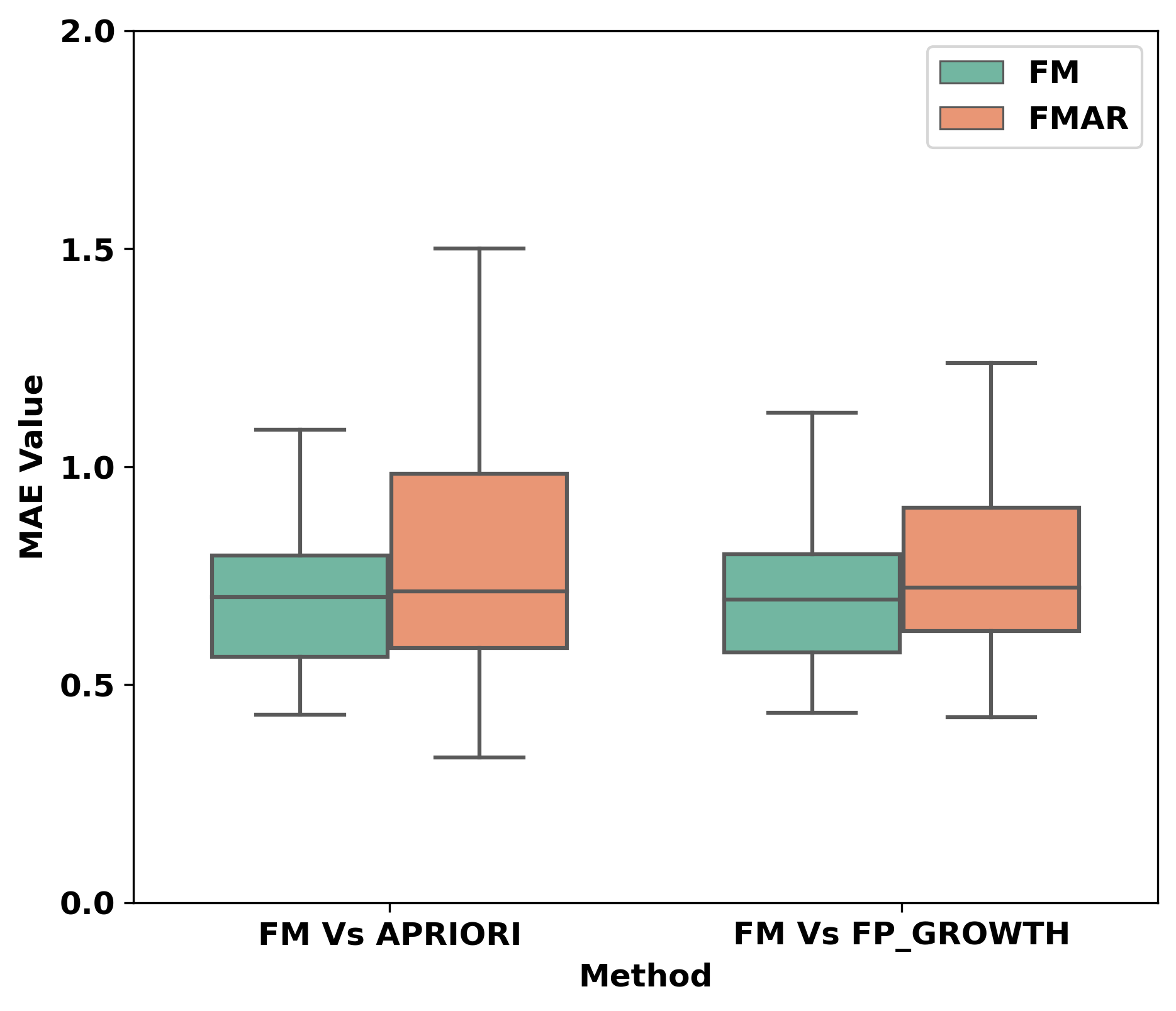}
  \captionof{figure}{MAE Comparison}
  \label{fig:mae_final}  
\end{minipage}%
\begin{minipage}{.5\textwidth}
  \centering
  \captionsetup{justification=centering}
  \includegraphics[width=1\linewidth,height=4.0cm]{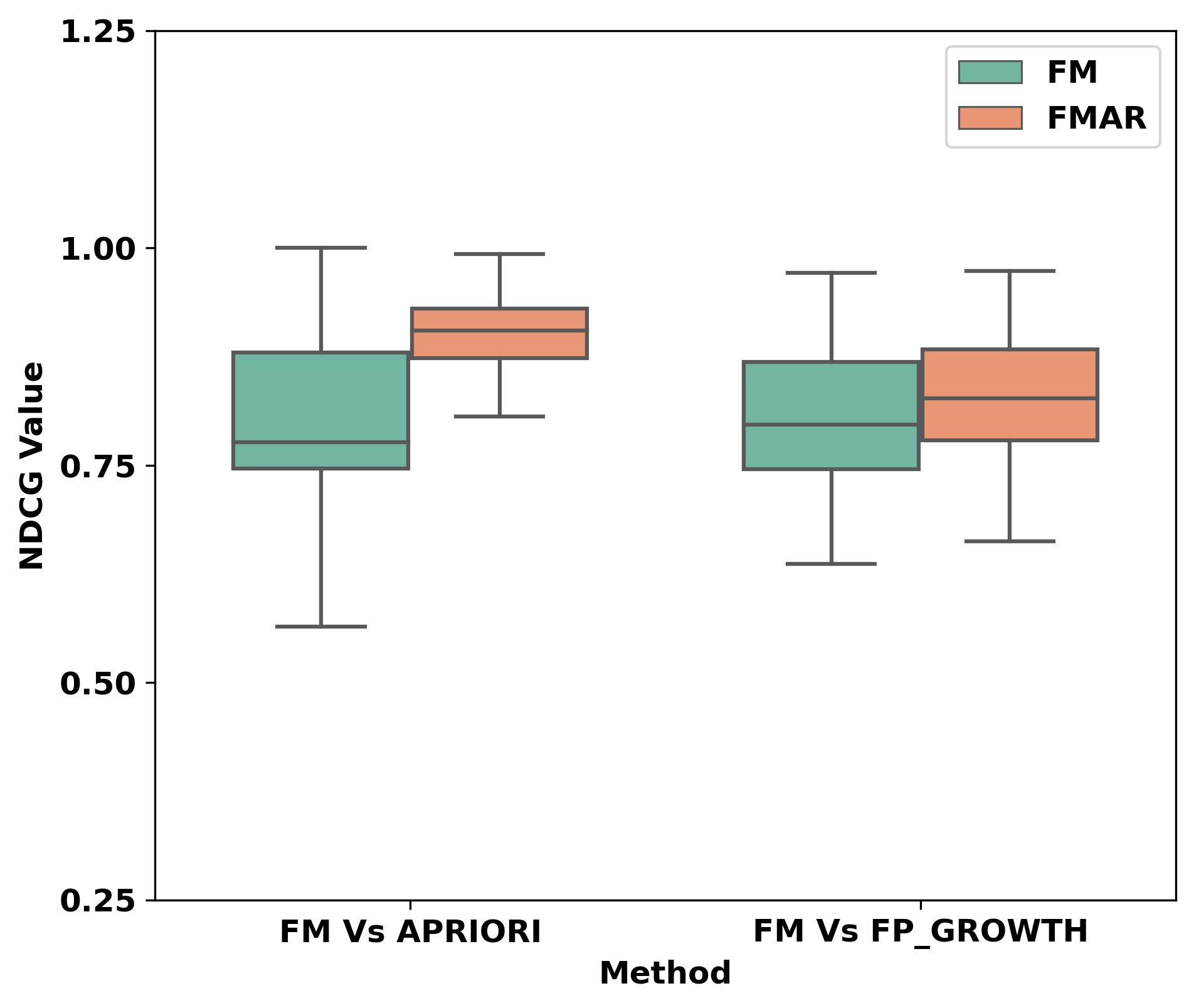}
  \captionof{figure}{NDCG Comparison}  
  \label{fig:ndcg_final}
\end{minipage}
\end{figure}

\indent In the first experiment, we calculate the mean absolute error (MAE) generated in both recommendation engines. The main goal of this approach is to show that the quality of recommendations is not significantly impacted after filtering the items in the testing set using the association rules. Fig~\ref{fig:mae_final} compares the mean absolute error of the predictions made using FM model with FM Apriori model and FM FP-growth model for 50 users. We use a box plot, which is a standardized way of displaying the distribution of data, to plot how the values of mean absolute error are spread out for 50 users. This graph encodes five characteristics of the distribution of data which show their position and length. These characteristics are minimum, first quartile (Q1), median, third quartile (Q3), and maximum. The results of this experiment show that MAE of FMAR in both versions, FM Apriori and FM FP-growth, is very close to MAE of FM recommender system. However, the average value of MAE for 50 users is $0.71$ using FM model, $0.80$ using FMAR (Apriori model), and $0.78$ using FMAR (FP Growth model).

% \begin{figure}
% \centering
% % \begin{minipage}{.5\textwidth}
%   \centering
%   \captionsetup{justification=centering}
%   \includegraphics[width=0.6\linewidth,height=5.0cm]{mae_final.png}
%   \captionof{figure}{MAE Comparison}
%   \label{fig:mae_final}  
% % \end{minipage}%
% % \begin{minipage}{.5\textwidth}
% %   \centering
% %   \captionsetup{justification=centering}
% %   \includegraphics[width=1\linewidth,height=3.5cm]{avg_prediction_fp_growth.png}
% %   \captionof{figure}{Average Prediction For Top 10 Items in FM FP-growth}  
% %   \label{fig:avg_pred_fp_growth}
% % \end{minipage}
% \end{figure}

% The slight difference between Apriori model and FP Growth model ca be interpreted due to the fact that the number of association rules generated using Apriori model is smaller than the number of association rules generated using FP Growth model.

% and how the values are spread out

% \indent In the first experiment, we calculate the average of prediction for the top 10 items which are generated in both recommendation engines. The main goal of this approach is to make sure that accuracy of predicted items is not highly impacted after filtering the items using the association rules. Fig~\ref{fig:avg_pred_apriori} and Fig~\ref{fig:avg_pred_fp_growth} compare the accuracy of FM model with FM Apriori model and FM FP-growth model for 50 users which are arbitrary chosen. The results proves that the accuracy of FMAR is very close to the accuracy of FM recommender system.

\indent In the second experiment, we evaluate FMAR by comparing its recommendation with FM using Normalized Discounted Cumulative Gain (NDCG) which is a measure of ranking quality that is often used to measure the effectiveness of recommendation systems or web search engines. NDCG is based on the assumption that highly relevant recommendations are more useful when appearing earlier in the recommendations list. So, the main idea of NDCG is to penalize highly relevant recommendations that appear lower in the recommendation list by reducing the graded relevance value logarithmically proportional to the position of the result. Fig~\ref{fig:ndcg_final} compares the accuracy of FM model with FM Apriori model and FM FP-growth model for 50 users and shows the distribution of the results. It is worth noting that in this test we calculate NDCG using the highest 10 scores in the ranking which are generated by FM or FMAR. The results show that both versions of FMAR model have always higher NDCG values than FM model which means that true labels are ranked higher by predicted values in FMAR model than in FM model for the top 10 scores in the ranking.

In the third evaluation method, we run Wilcoxon Rank-Sum test on the results of previous experiments. Firstly, we apply Wilcoxon Rank-Sum test to the results of the first experiment in order to prove that the difference in MAE between FM and FMAR is not significant, and hence, it can be discarded. So, we pass two sets of samples of MAE for FM and FMAR. The Table~\ref{table:wilcoxon_test} shows the p-value for comparing FMAR using Apriori model and FP Growth model with FM model. In both cases, we got p-value $>$ $0.05$ which means that the null hypothesis is accepted at the $5\%$ significance level, and hence, the difference between the two sets of measurements is not significant. On the other hand, we apply Wilcoxon Rank-Sum test to the results of the second experiment to check if the difference in NDCG between FM and FMAR is significant. However, the Table~\ref{table:wilcoxon_test} shows that p-value $<$ $0.05$ for comparing both models of FMAR with FM model. This means the null hypothesis is rejected at the $5\%$ significance level (accept the alternative hypothesis), and hence, the difference is significant. Since FMAR model has higher NDCG than FM model, we can conclude that FMAR outperforms FM for the highest top 10 predictions.

% are drawn from the same distribution

\renewcommand{\arraystretch}{1.5}
\begin{table}
\centering
\begin{tabular}{M{6.1cm}||M{4.5cm}} 
 \hline
 Model & p-value \\ 
 \hline\hline
 MAE (FM Vs FMAR-Apriori model) & 0.29 \\ 
 \hline
 MAE (FM Vs FMAR-FP Growth model) & 0.13 \\
 \hline\hline
 NDCG (FM Vs FMAR-Apriori model) & 1.74e-08 \\
 \hline
 NDCG (FM Vs FMAR-FP Growth model) & 0.04 \\
 \hline
\end{tabular}
\renewcommand{\arraystretch}{1}
\caption{Wilcoxon Rank-Sum Test}
\label{table:wilcoxon_test}
\end{table}

% \begin{table*}
% \caption{Wilcoxon Rank-Sum test}
% \label{6}
% \begin{tabularx}{\linewidth}{c|c}
% \cmidrule{1-2}
% Model &
% p\_value \\
% \cmidrule{1-2}
% % \multicolumn{23}{c}{\makecell{RMSE \\ MAE }} \\
% \cmidrule{1-2}
% MAE (Apriori model) &
% \makecell{0.29}   \\ 
% % \addlinespace
% \cmidrule{1-2}
% MAE (FP Growth model) &  
% \makecell{0.19}  \\ 
% \cmidrule{1-2}
% \end{tabularx}
% \end{table*}

%In the last experiment, we compare between FM and FMAR in terms of the number of the items that we need to make prediction for. The main idea here is to estimate how much the elapsed time necessary to make a prediction for any user is decreased in FMAR. Fig~\ref{fig:num_pred_final} shows the distribution of the results of this experiment for 50 users. It is needless to say that the number of items that we need to predict for in FMAR has been significantly decreased after using the association rules. Moreover, the results show that FMAR model can generate predictions for any user at least 4 times faster than FM model. Finally, it is noteworthy that there is no need to regenerate or update the association rules frequently in FMAR. Therefore, the computational cost of extracting the association rules is not considered in our comparisons.

\begin{figure}[t]
\centering
\includegraphics[width=10cm, height=5cm]{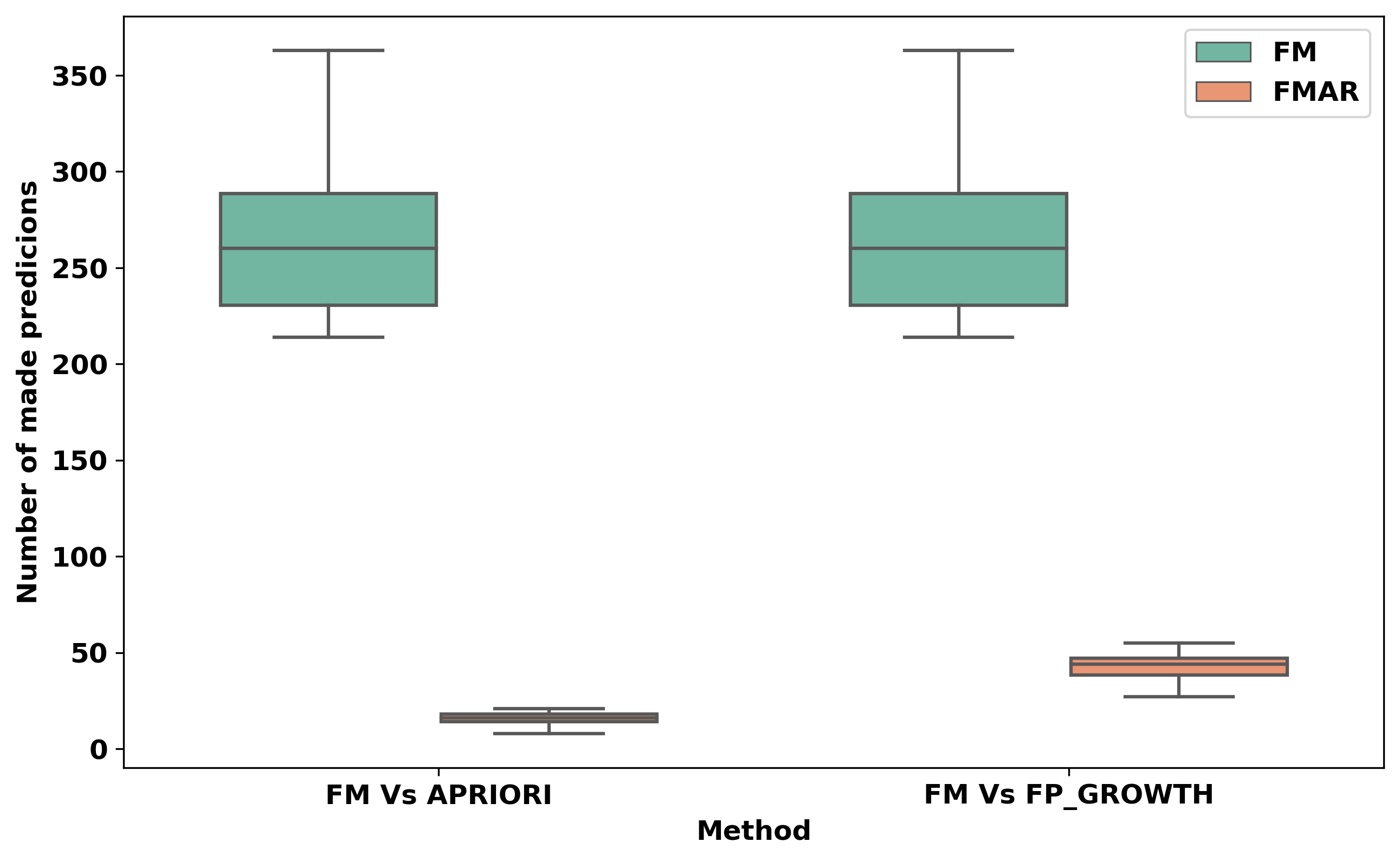}
\caption{Comparison of the speed of methods (estimated by the number of predictions made by the factorization machine model, i.e., the lower the better)}
\label{fig:num_pred_final}
\end{figure}

In the last experiment, we compare FM and FMAR in terms of the speed of their operation, measured as the number of predictions performed by the factorization machines model. The main idea is to estimate the time necessary to prepare recommendations for every tested user for both evaluated approaches. Fig~\ref{fig:num_pred_final} shows the distribution of the results of this experiment for the selected 50 test users. Observably, the number of items that we need to predict with FMAR is significantly lower due to using the association rules for filtering. The results show that the FMAR model can generate predictions for any user at least four times faster than the FM model. Finally, it is noteworthy that generating the rules is part of the training procedure. Therefore, it is a one-time effort, and there is no need to regenerate or update the association rules frequently in FMAR. Therefore, the computational cost of the training procedure for every method, including extracting the association rules, is not considered in our comparisons. 

In the final analysis, all previous experimentations showed that after applying our method, the factorization machines could perform significantly faster with no drop in quality considering MAE and NDCG measures.

% In the final analysis,
% In a nutshell,
% As a conclusion,
% in summary,

% \vspace{-1.0em}

\section{Conclusions and Future Work \label{sec:summary}}

This article introduces FMAR, a novel recommender system, which methodically incorporates the association rules in generating the recommendations using the factorization machines model. Our study evaluates two approaches to creating association rules based on the users' rating history, namely: the apriori and frequent pattern growth algorithms. These rules are used to decrease the number of items passed to the model to estimate the ratings, reducing the latency of the recommender system prediction.

%In this paper, we have proposed FMAR, a novel recommender system, which methodically incorporates the association rules in the process of generating the recommendations using factorization machines model.
% The main goal of this paper is to improve the efficiency of the recommender systems rather than accuracy which has been heavily studied and improved in the past few years.
%We suggested to use apriori algorithm or frequent pattern (FP) growth algorithm to create a set of association rules based on the rating history. These rules are used to decrease the number of items which are passed to factorization machines model to estimate the ratings, and hence, decrease the prediction latency of the recommender system.

To evaluate our proposed model, we conducted comprehensive experiments on MovieLens 100K dataset using the libFM tool \cite{c12} which provides implementations for factorization machines as well as machine learning algorithms. Moreover, we presented multiple evaluation methods to compare the performance of FMAR against the recommender system built using the factorization machines algorithm. The experimental results show that FMAR has improved the efficiency of recommender systems. Furthermore, the experiments also indicate that the accuracy of FMAR is very close to the results produced by the standard recommender system.

%In the future work, we plan to incorporate more information in the process of producing the association rules, such as users' demographics, items' characteristics and contextual information. Also, we are interested in creating a web interface where FMAR is used to generate recommendations for users. In this scenario, we are particularly interested in employing more advanced algorithms to generate the association rules, such as AprioriTID \cite{c4}\cite{c6}, Apriori Hybrid \cite{c4}\cite{c6}, AIS (Artificial Immune System) \cite{c7} and SETM \cite{c8}. We believe that using previous algorithms would help to further improve the performance and accuracy of FMAR. Furthermore, a set of datasets with different sizes will be available in the suggested tool. As a result, FMAR can be evaluated using different settings which can be selected in a user-friendly web-interface. On the other hand, we need to take into account the dynamics of evolving the association rules by periodically updating them based on recent changes in rating history. Another direction of future work is to utilize the generated association rules to solve the cold-start problem in recommendation systems where the new users do not have, or have very few, ratings in the past. Finally, we plan to use distributed stream processing engines, like Apache Flink, to examine parallel implementations of FMAR, where the process of extracting the rules and generating the recommendations is scalable to infinite streams or large-scale datasets. 

In the future work, we plan to incorporate more information in the process of producing the association rules, such as users' demographics, items' characteristics, and contextual information. Another important aspect to consider is to evaluate our proposed model using different recommender systems and different sizes of datasets. However, we are interested in creating a web interface where FMAR is used to generate recommendations for users. In this scenario, we are particularly interested in employing more advanced algorithms to generate the association rules, such as AprioriTID \cite{c4}\cite{c6}, Apriori Hybrid \cite{c4}\cite{c6}, AIS (Artificial Immune System) \cite{c7} and SETM \cite{c8}. We believe that using previous algorithms would help to further improve the performance and accuracy of FMAR. As a result, FMAR can be evaluated using different settings selected in a user-friendly web interface.

Furthermore, we plan to consider the changes in users' behavior and preferences by periodically updating association rules based on recent changes in rating history. Another direction of future work is to utilize the generated association rules to solve the cold-start problem in recommendation systems where the new users do not have (or have very few) ratings in the past. Finally, we plan to use distributed stream processing engines, like Apache Flink, to examine parallel implementations of FMAR, where the process of extracting the rules and generating the recommendations is scalable to vast streams or large-scale datasets. 

\end{document}